\documentclass[final]{cvpr}
\usepackage{times}
\usepackage{epsfig}
\usepackage{graphicx}
\usepackage{amsmath}
\usepackage{amssymb}

\usepackage{multirow}
\usepackage{tabularx}
\usepackage{booktabs}
\usepackage{caption}
\usepackage{bm}
\usepackage[ruled,vlined]{algorithm2e}
\usepackage{algpseudocode}
\usepackage{colortbl}
\definecolor{grayDark}{gray}{0.95}
\definecolor{grayLight}{gray}{0.98}

\newcommand{\st}[1]{\textcolor{red}{\textbf{#1}}}
\newcommand{\nd}[1]{\textcolor{blue}{\textbf{#1}}}
\newcommand{\te}[2]{$\text{#1}_{\uparrow \text{#2}\%}$} 

\usepackage[pagebackref=true,breaklinks=true,colorlinks,bookmarks=false]{hyperref}

\def\cvprPaperID{1796} 
\def\confYear{CVPR 2021}

\title{Exploring Intermediate Representation for Monocular Vehicle Pose Estimation}
\author{Shichao Li\textsuperscript{1},\quad Zengqiang Yan\textsuperscript{1},\quad Hongyang Li\textsuperscript{2},\quad Kwang-Ting Cheng\textsuperscript{1}\\
	\textsuperscript{1}The Hong Kong University of Science and Technology,\quad  \textsuperscript{2}SenseTime Research \\ {\tt\small slicd@cse.ust.hk},
	\quad{\tt\small timcheng@ust.hk}
}
\begin{document}
\twocolumn[{%
\renewcommand\twocolumn[1][]{#1}%
\maketitle
\begin{center}
	\centering
	\includegraphics[width=1\linewidth, trim=0cm 1.5cm 1cm 4cm]{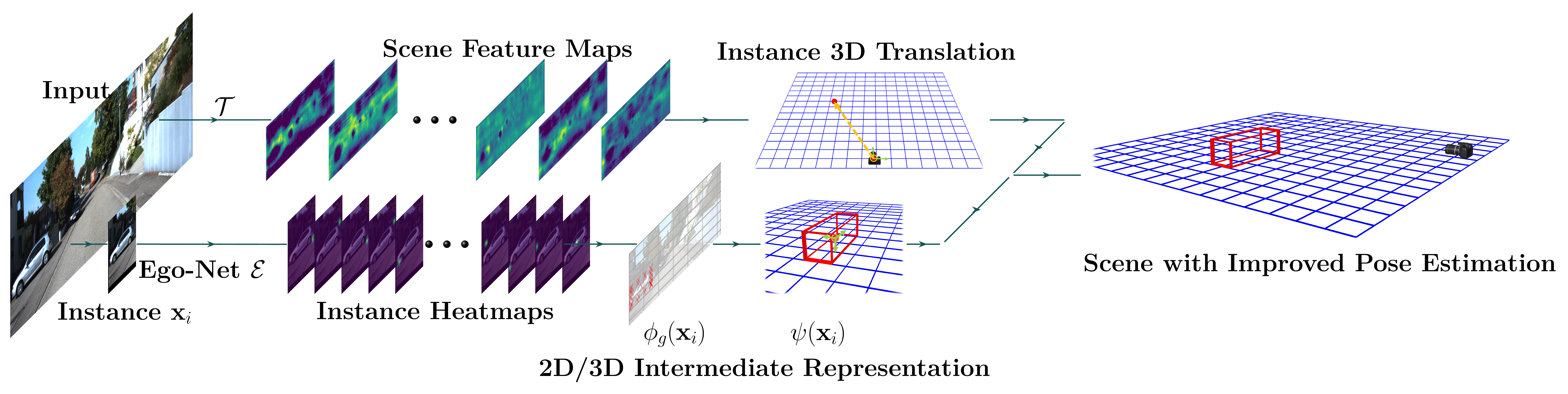}
	\captionof{figure}{A novel deep model, Ego-Net, is proposed for accurate vehicle orientation estimation in the camera coordinate system via extracting a set of explicit 2D/3D geometrical representations. A 3D vehicle detection system (upper branch) can enjoy improved orientation estimation accuracy by replacing its pose estimation module with Ego-Net (lower branch).}
	\label{architecture}
\end{center}%
}]

\begin{abstract}  
We present a new learning-based framework to recover vehicle pose in SO(3) from a single RGB image. In contrast to previous works that map local appearance to observation angles, we explore a progressive approach by extracting meaningful Intermediate Geometrical Representations (IGRs) to estimate \textbf{egocentric} vehicle orientation. This approach features a deep model that transforms perceived intensities to IGRs, which are mapped to a 3D representation encoding object orientation in the camera coordinate system. Core problems are \textbf{what} IGRs to use and \textbf{how} to learn them more effectively. We answer the former question by designing IGRs based on an interpolated cuboid that derives from primitive 3D annotation readily. The latter question motivates us to incorporate geometry knowledge with a new loss function based on a projective invariant. This loss function allows unlabeled data to be used in the training stage to improve representation learning. Without additional labels, our system outperforms previous monocular RGB-based methods for joint vehicle detection and pose estimation on the KITTI benchmark, achieving performance even comparable to stereo methods. Code and pre-trained models are available at this https URL\footnote{\url{https://github.com/Nicholasli1995/EgoNet}}. {\let\thefootnote\relax\footnote{{Part of this work was done when the first author was an intern at SenseTime Research Shanghai Autonomous Driving Group.}}}
\end{abstract}
\begin{figure*}[t]
	\begin{center}
		\includegraphics[width=0.9\linewidth, trim=4cm 2cm 13cm 2cm]{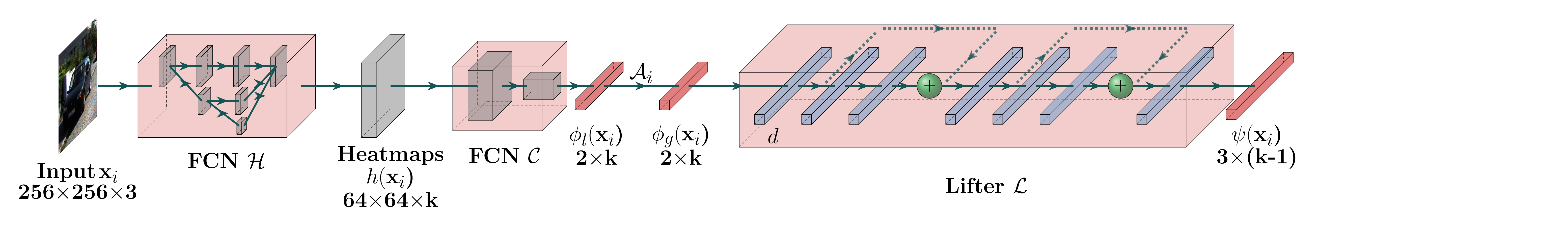}
	\end{center}
	\caption{Detailed architecture of Ego-Net. Feature maps are first computed with a fully convolution model $\mathcal{H}$ from a detected instance. Heatmaps representing the projection of a 3D cuboid are regressed and mapped to local coordinates with several strided convolution layers. The local coordinates are transformed to screen coordinates $\phi_g(\mathbf{x}_i)$ and mapped to a 3D cuboid representation $\psi(\mathbf{x}_i)$, whose orientation directly represent egocentric pose in the camera coordinate system. k=33 when q=2 as in Sec. \ref{MA}.}
	\label{fig:egonet}
\end{figure*}
\section{Introduction}
\begin{quote}
\textit{``The usefulness of a representation depends upon how well suited it is to the purpose for which it is used".}
\hfill
--David Marr~\cite{marr2010vision}
\end{quote}
Understanding 3D properties of the surrounding world is critical for vision-based autonomous driving and traffic surveillance systems~\cite{ferryman2000visual}. Accurate 3D vehicle orientation estimation (VOE) can imply a driver's intent of travel direction, help predicting its position a moment later and help identifying anomalous behaviors. To pursue high accuracy, recent studies have exploited active sensors to obtain point clouds~\cite{yang2018pixor, zhou2018voxelnet, qi2019deep} or the use of stereo cameras~\cite{li2018stereo, sun2020disp, peng2020ida}. The former enjoys the convenience of directly working with 3D data and the latter benefit from multi-view geometry. As downsides, LiDAR sensors cost more than commodity RGB cameras and stereo systems incur extra complexity. Pushing the limit of a single-view RGB-based method is not only cost-effective, but can also complement other configurations through sensor fusion.  

However, estimating 3D parameters from a single RGB image is highly challenging due to the lack of depth information and other geometrical cues. Thanks to the strong representation learning power brought by recent advances in deep architectures, state-of-the-art (SOTA) RGB-based methods can regress 3D parameters from visual appearance in an end-to-end manner~\cite{mousavian20173d, brazil2019m3d, simonelli2019disentangling}. In this paradigm, paired images and 3D annotations are specified as inputs and learning targets respectively for supervising a deep model. No intermediate representation is designed in such end-to-end solutions, which need a large amount of training pairs to approximate the highly non-linear mapping from the pixel space to 3D geometrical quantities.

In contrast to the aforementioned methods, we take a different approach and ask \emph{can a neural network extract explicit geometrical quantities from RGB images and use them for monocular vehicle pose estimation}? Our inspiration is from the representational framework for vision~\cite{marr2010vision}, which first extracts geometrically meaningful $2\frac{1}{2}$-D sketch from perceived intensities and converts them to 3D representations. We are also inspired by recent advances in two-stage 3D human pose estimation~\cite{li2020cascaded}. Previous approaches~\cite{mousavian20173d, brazil2019m3d, simonelli2019disentangling} specify the start (pixels) and the end (3D pose) of such a process while ignoring the learning of any intermediate representation. Apart from the motivation from the framework~\cite{marr2010vision}, geometrical representations have two other benefits: (1)~they are easier for human to comprehend than the dense feature maps when debugging safety-critical applications, (2)~the mapping from them to 3D pose is easy to learn as a few layers would suffice in our experiments.   

Towards this end, we propose a novel VOE approach featuring learning Intermediate Geometrical Representations (IGRs). We first explicitly define a set of IGRs based on existing ground truth without any extra manual annotation. We then design a novel deep model, Ego-Net, which first regresses IGRs and then maps them to the 3D space,  achieving SOTA performance. Apart from the high accuracy, we propose a novel loss function for self-supervising the learning of IGRs based on prior knowledge of projective invariants. With a hybrid source of supervision, our model can benefit from unlabeled data for enhanced generalization. Our contributions can be summarized in the following:   
\begin{itemize}
\item To our best knowledge, our proposed learning-based model, Ego-Net, is the first one which regresses intermediate geometrical representations  for \emph{direct} estimation of egocentric vehicle pose.
\item Our novel loss function based on the preservation of projective invariants introduces self-supervision for the learning of such representations.
\item Our approach does not require additional labels, and outperforms previous monocular RGB-based VOE methods. Our monocular system achieves performance comparable to some stereo methods for joint vehicle detection and pose estimation on KITTI.	
\item Ego-Net, a plug-and-play module, can be incorporated into any existing 3D object detection models for improved VOE. Equipped with Ego-Net, a 3D detection system can achieve high 3D location and orientation estimation performance simultaneously, while previous works~\cite{brazil2019m3d, Ding_2020_CVPR, kinematic-3d} usually fail to achieve both.
\end{itemize}

After discussing relevant works in Sec.~\ref{RW}, we detail the design of Ego-Net in Sec.~\ref{MA}. In Sec.~\ref{PI} we demonstrate how unlabeled data can be utilized to learn IGRs with prior geometry knowledge. Sec.~\ref{EX} presents experimental results. 

\section {Related Works}
\label{RW}
\noindent \textbf{Deep learning-based 3D object detection} aims to learn a mapping from sensor input to instance attributes such as 3D orientation and 2D/3D location. Depending on the sensor modality, recent works can be categorized into LiDAR-based approaches~\cite{zhou2018voxelnet, lang2019pointpillars, shi2019pointrcnn, yang2019std, zhao20193d, shi2020point, meng2020weakly}, RGB-based methods~\cite{mousavian20173d, xu2018multi, roddick2019orthographic, simonelli2019disentangling, li2019stereo, cai2020monocular}, and hybrid ones~\cite{xu2018pointfusion, vora2020pointpainting}. Our method falls into the second category, which does not require expensive devices to acquire point clouds. Compared to stereo approaches~\cite{li2018stereo, li2019stereo, xu2020zoomnet, chen2020dsgn} that need two RGB cameras, our system only takes a \emph{single view} input. For convenience in implementation, many previous works~\cite{kundu20183d, brazil2019m3d, Ding_2020_CVPR, liu2020smoke, Ma_2020_ECCV, shi2020distance} employ a multi-task architecture composed of a backbone for feature extraction and 3D parameter estimation heads. Each head shares similar design and VOE is treated as a sub-task with equal importance as other tasks such as dimension estimation. In contrast, we present a dedicated module for this task which outperforms previous ones.   

\noindent \textbf{Pose estimation in 3D object detection systems} targets at recovering instance orientation. We focus on the works that use RGB images and classify them into \emph{learning-based} methods and geometrical \emph{optimization-based} ones. 

\emph{Learning-based} methods learn a function from pixels to orientation. Conventionally, visual features were extracted and fed to a model for discrete pose classification (DPC) or continuous regression. Early works~\cite{juranek2015real, xiang2015data} utilized hand-crafted features~\cite{dollar2014fast} and boosted trees for DPC. Recent works replace the feature extraction stage with pre-designed deep models. ANN~\cite{yang2014object} and DAVE~\cite{zhou2016dave, zhou2017fast} classify instance feature maps extracted by CNN into discrete bins. To deal with images containing multiple instances, Fast-RCNN-like architectures were employed in~\cite{chen20153d, braun2016pose, chen20173d, huang2019perspectivenet, Ke_2020_ECCV} where region-of-interst (ROI) features were used to represent instance appearance and a classification head gives pose prediction. Deep3DBox~\cite{mousavian20173d} proposed \emph{MultiBin} loss for joint pose classification and residual regression. Wasserstein loss was promoted in~\cite{liu2019conservative} for DPC. LSTM was employed in~\cite{hu2019joint} to refine vehicle pose using temporal information.
Our work is also a learning-based approach but possesses key differences compared to previous works. Our approach promotes learning explicit IGRs while previous works do not. With IGRs, later part of our model can directly reason in the geometrical space. Such design enables our model to directly estimate global (egocentric) pose in the camera coordinate system while previous works can only estimate relative (allocentric) pose and require a second step to obtain the global pose.  

\emph{Optimization-based} methods first extract 2D evidence from RGB images and fit a 3D model to the evidence for pose estimation. 
Deepmanta~\cite{chabot2017deep} detected pre-annotated semantic object parts and used Epnp~\cite{lepetit2009epnp} for pose optimization. Mono3D++~\cite{he2019mono3d++} introduced task priors into a joint energy function and conducted optimization using the Ceres solver~\cite{ceres-solver}. Compared to these works, our approach does not need the construction of a 3D shape model neither any extra manual labeling. Such labeling was required in~\cite{chabot2017deep} to train a model for detecting image evidence. Neither does our approach need initialization and iterative optimization for inference. In fact, our approach can serve as the initialization step and be integrated with these works.  

\noindent \textbf{Self-supervision from geometry knowledge} can be incorporated into deep systems to regularize representation learning. Epipolar geometry was used in~\cite{kocabas2019self} to generate supervisory signal for a fully convolutional network. Loop consistency~\cite{chen2019unsupervised} and projective consistency~\cite{habibie2019wild} were employed for 3D human pose estimation. Flow fields and disparity maps were extracted in~\cite{gan2018geometry} for self-supervised video representation learning. We propose a new self-supervision mechanism by utilizing projective invariants to supervise our intermediate representations. This approach can utilize unlabeled images during training yet does not require multi-view setting as~\cite{kocabas2019self}. It is also camera-agnostic and does not require camera intrinsics as~\cite{habibie2019wild}.


\section{Egocentric 3D Vehicle Pose Estimation}
\label{MA}
Ego-Net $\mathcal{E}$ estimates 3D orientation from cropped instance appearance. It can be combined with a 3D location estimation module $\mathcal{T}$ to form a 3D object detection model $\mathcal{M} = (\mathcal{E}, \mathcal{T})$ as shown in Fig.~\ref{architecture}. We detail $\mathcal{E}$ in the following and some examples of composing $\mathcal{M}$ are in our supplementary material. 

\subsection{Single-step Progressive Mapping}
\label{Ego-Net}
\begin{figure}[h]
	\begin{center}
		\includegraphics[width=0.95\linewidth, trim=4cm 5cm 4cm 5cm]{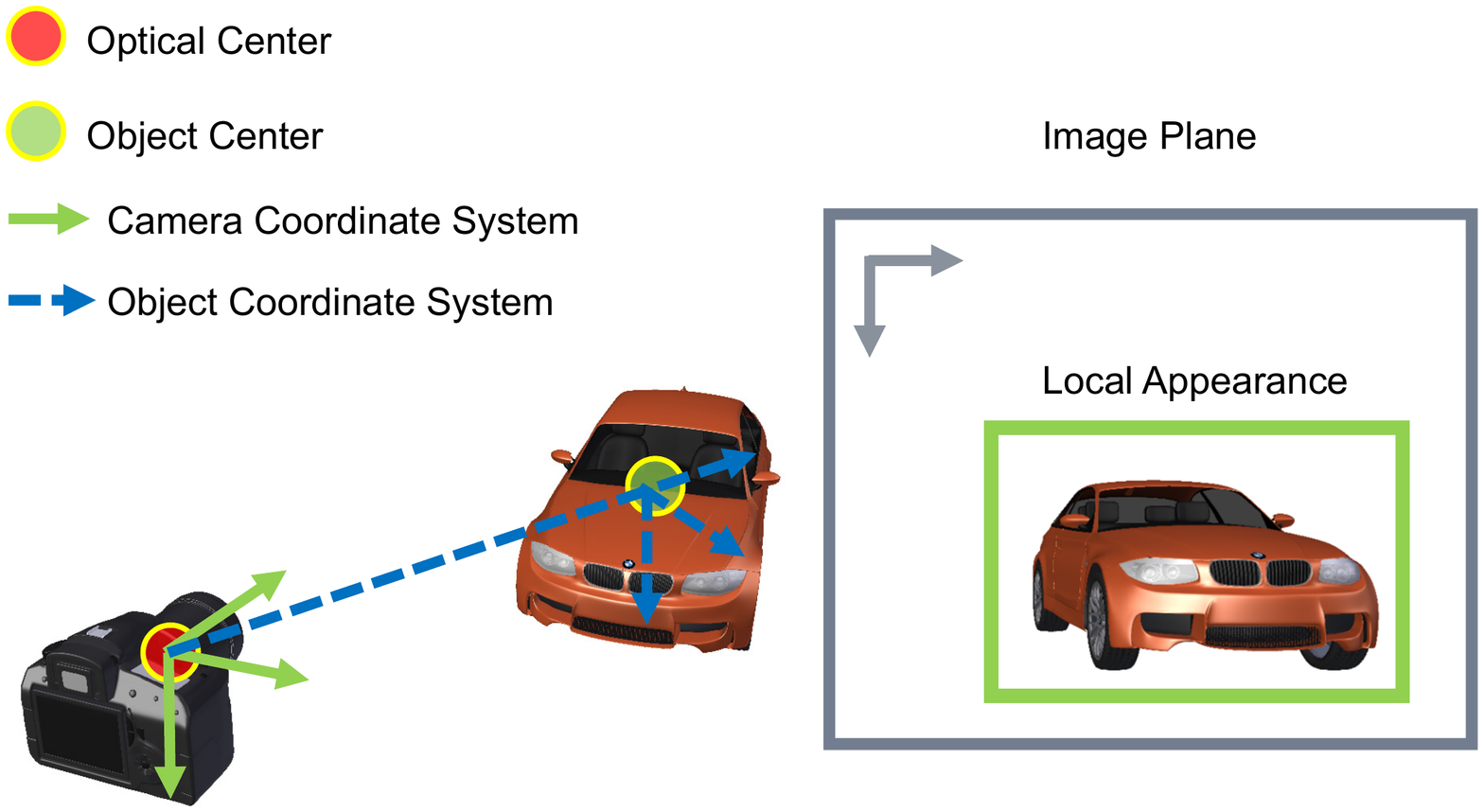}
	\end{center}
	\caption{Local appearance cannot uniquely determine egocentric pose. Existing solutions first estimate allocentric pose in the object coordinate system (blue) and convert it to egocentric pose in the camera coordinate system (green) based on the object location.}
	\label{fig:3.1}
\end{figure}

For visual perception and traffic surveillance systems, expressing the poses of multiple objects in the \emph{same} coordinate system is favorable and convenient for downstream planing and decision-making. Previous works~\cite{mousavian20173d, kundu20183d, brazil2019m3d, Ding_2020_CVPR} achieved this by adopting the computational graph in Eq.~\ref{eq:comp_graph_previous}. A CNN-based model $\mathcal{N}$ is used to map local instance appearance $\mathbf{x}_i$ to allocentric pose, i.e., 3D orientation in the object coordinate system $\{\mathbf{i}, \mathbf{j}, \mathbf{k}\}$, which is later converted to the egocentric pose, i.e., orientation in the camera coordinate system. The difference between these two coordinates systems is shown in Fig.~\ref{fig:3.1}. This two-step design is a workaround since an object with the same egocentric pose $\bm{\theta}_i$ can produce different local appearance depending on its location~\cite{kundu20183d} and learning the mapping from $\mathbf{x}_i$ to $\bm{\theta}_i$ is an ill-posed problem. In this two-step design, $\{\mathbf{i}, \mathbf{j}, \mathbf{k}\}$ needs to be estimated by another parallel module and $\mathcal{N}$ \emph{can not recover $\bm{\theta}_i$ independently} without the assist of this module. The error of this module can propagate to the final estimation of egocentric poses. Even with this workaround, the problem remains challenging as the mapping from $\mathbf{x}_i$ to $\bm{\alpha}_i$ is highly non-linear and difficult to approximate.
\begin{equation}
\label{eq:comp_graph_previous}
\begin{array}{llll}
\mathbf{x}_i  & \xrightarrow{\mathcal{N}} & ~~ \bm{\alpha}_i & \xrightarrow{convert} ~~ \bm{\theta}_i \\
& & \{\mathbf{i}, \mathbf{j}, \mathbf{k}\} & \nearrow{assist}\\
\end{array}
\end{equation}

We believe that such two-step computation is unnecessary and propose to learn a direct mapping $\mathcal{E}$ from visual information to egocentric pose. However, instead of relying on a black box model to fit such a non-linear mapping, we assume this mapping to be a progressive process, where a series of IGRs are gradually extracted from pixels and eventually lifted to the 3D target. Specifically, we design Ego-Net as a composite function consisting of $\{\mathcal{H},\mathcal{C}, \mathcal{L}\}$ described as follows. Given the cropped patch of one detected instance $\mathbf{x}_i$, Ego-Net predicts its egocentric pose as $\mathcal{E}(\mathbf{x}_i) = \mathcal{L}(\mathcal{A}_i(\mathcal{C}(\mathcal{H}(\mathbf{x}_i)))) =  \bm{\theta}_i$. Fig.~\ref{fig:egonet} depicts Ego-Net, whose computational graph is shown in Eq.~\ref{eq:comp_graph}. $\mathcal{H}$ is assumed to extract heatmaps $h(\mathbf{x}_i)$ for salient 2D object parts that are mapped by $\mathcal{C}$ to coordinates $\phi_l(\mathbf{x}_i)$ representing their local location within the patch. $\phi_l(\mathbf{x}_i)$ is converted to the global image plane coordinates $\phi_g(\mathbf{x}_i)$ with an affine transformation $\mathcal{A}_i$ parametrized with scaling and 2D translation. $\phi_g(\mathbf{x}_i)$ is further lifted into a 3D representation $\psi(\mathbf{x}_i)$ by $\mathcal{L}$. The final pose prediction derives from $\psi(\mathbf{x}_i)$.        
\begin{equation}
\label{eq:comp_graph}
\arraycolsep=1pt
\begin{array}{lllllllllll}
\mathbf{x}_i  & \xrightarrow{\mathcal{H}} & h(\mathbf{x}_i) & \xrightarrow{\mathcal{C}} & \phi_l(\mathbf{x}_i) & \xrightarrow{\mathcal{A}_i} & \phi_g(\mathbf{x}_i) & \xrightarrow{\mathcal{L}} & \psi(\mathbf{x}_i)  & \rightarrow & \bm{\theta}_i
\end{array}
\end{equation} 

\subsection{Labor-free Intermediate Representations}
The abstraction in Eq.~\ref{eq:comp_graph} does not mathematically define what geometrical quantity to use as IGRs, here we define them based on three considerations:

\noindent \emph{Availability}: The IGRs should be easily derived from existing ground truth annotations with none or minimum extra manual effort. While using CAD models such as the wire-frame model in~\cite{chabot2017deep} and 3D voxel pattern in ~\cite{xiang2015data} can provide high level-of-detail, such representations require extensive manual effort and are not scalable. A new driving scenario may requires collecting a new training set, and can incur significant burden if the IGRs cannot be readily obtained. In addition, no evidence has shown that detailed object shapes are necessary for orientation estimation.

\noindent \emph{Discriminative}: The IGRs should be indicative for 3D pose estimation, so that it can serve as a good bridge between visual appearance input and the geometrical target.

\noindent \emph{Transparency}: The IGRs should be easy to understand, which makes them debugging-friendly and trustworthy for applications such as autonomous driving. While network visualization techniques~\cite{simonyan2013deep, li2019visualizing, li2019facial} can be used to understand learned representations, designing IGRs with explicit meaning offers more convenience in practice.

\begin{figure}[t]
	\begin{center}
		\includegraphics[width=0.5\linewidth, trim=8cm 5cm 8cm 3cm]{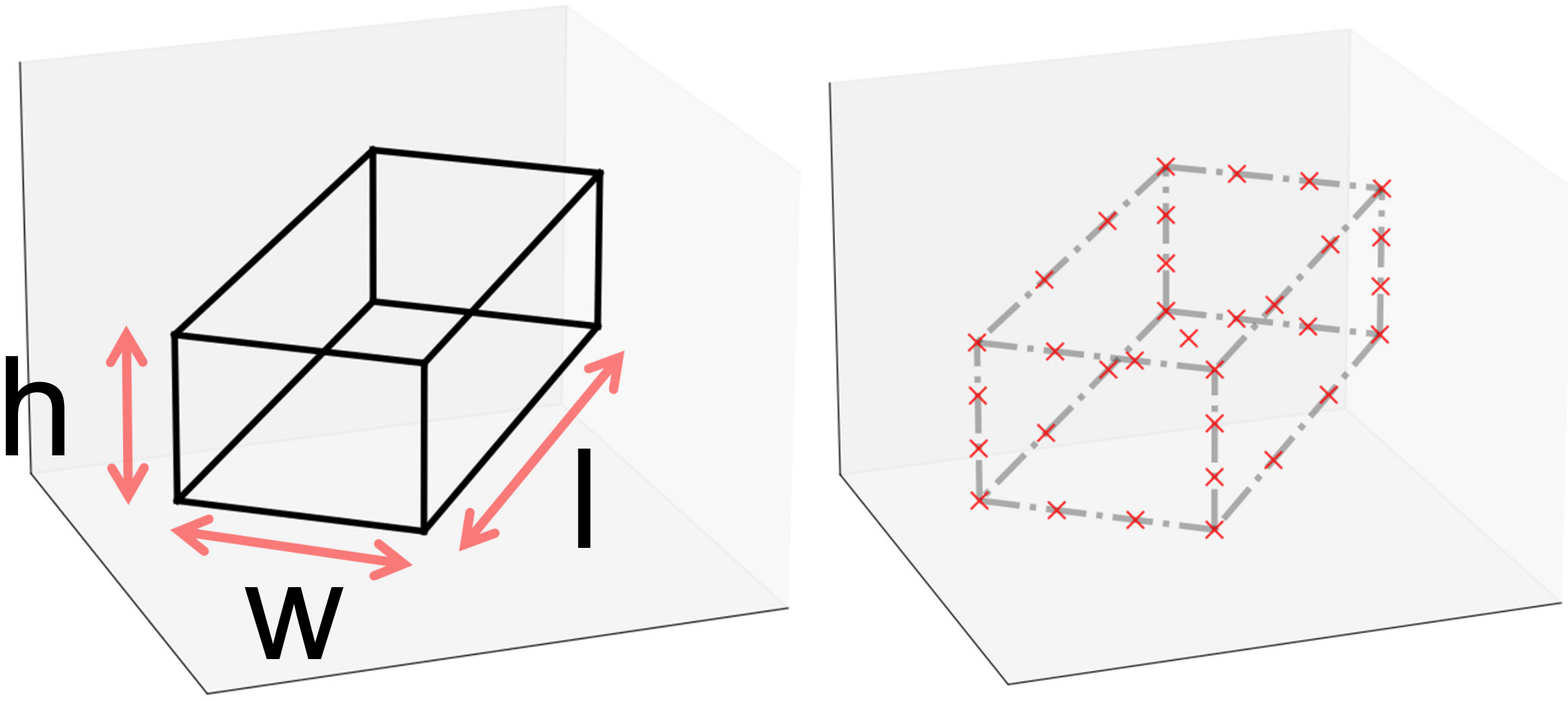}
	\end{center}
	\caption{An instance $\mathbf{x}_i$ with primitive 3D bounding box annotation (left) and its 3D point cloud representation $\psi(\mathbf{x}_i)$ (right) in the camera coordinate system as the collection of red crosses.}
	\label{fig:3drep}
\end{figure}

With the above considerations, we define the 3D representation  $\psi(\mathbf{x}_i)$ as sparse 3D point cloud representing an interpolated cuboid. Autonomous driving datasets such as KITTI~\cite{geiger2012we} usually label instance 3D bounding boxes from captured point clouds where an instance $\mathbf{x}_i$ is associated with its centroid location in the camera coordinate system $\mathbf{t}_i = [t_x, t_y, t_z]$, size $[h_i, w_i, l_i]$, and its egocentric pose $\bm{\theta}_i$. As shown in Fig.~\ref{fig:3drep}, denote the 12 lines enclosing the vehicle as $\{\mathbf{l}_j\}_{j=1}^{12}$, where each line is represented by two endpoints (\textbf{s}tart and \textbf{e}nd) as $\mathbf{l}_j = [\mathbf{p}_j^{s}; \mathbf{p}_j^{e}]$. $\mathbf{p}_j^v$  ($v$ is $s$ or $e$) is a 3-vector $(X_j^{v}, Y_j^{v}, Z_j^{v})$ representing the point's location in the camera coordinate system. As a complexity-controlling parameter, $q$ more points are derived from each line with a pre-defined interpolation matrix $B_{q \times 2}$ as
\begin{equation}
\begin{bmatrix}
\mathbf{p}_j^1 \\
\mathbf{p}_j^2 \\
\dots \\
\mathbf{p}_j^q \\
\end{bmatrix}
=
B_{q \times 2}
\begin{bmatrix}
\mathbf{p}_j^{s} \\
\mathbf{p}_j^{e} \\
\end{bmatrix}
=
\begin{bmatrix}
\beta_1 & 1 - \beta_1 \\
\beta_2 & 1 - \beta_2 \\
\dots & \dots \\
\beta_q & 1 - \beta_q \\
\end{bmatrix}
\begin{bmatrix}
\mathbf{p}_j^{s} \\
\mathbf{p}_j^{e} \\
\end{bmatrix}.
\end{equation} 
The 8 endpoints, the instance's centroid, and the interpolated points for each of the 12 lines form a set of $9 + 12q$ points. The concatenation of these points forms a $9 + 12q $ by $3$ matrix $\tau(\mathbf{x}_i)$. Since we do not need the 3D target $\psi(\mathbf{x}_i)$ to encode location, we deduct the instance translation $\mathbf{t}_i$ from $\tau(\mathbf{x}_i)$ and represent $\psi(\mathbf{x}_i)$ as a set of $8 + 12q$ points representing the shape relative to the centroid  
\begin{equation}
\psi(\mathbf{x}_i) = \{(X_{j}^{v} - t_x, Y_{j}^{v} - t_y, Z_{j}^{v} - t_z)\}
\end{equation}
where $v\in\{s,1,\dots,q,e\}$ and $j\in\{1,2,\dots,12\}$.
Larger $q$ provides more cues for inferring pose yet increases complexity. In practice we choose $q=2$ and the right figure of Fig.~\ref{fig:3drep} shows an example with 
$B_{2 \times 2}= \begin{bmatrix}
\frac{3}{4} & \frac{1}{4} \\
\frac{1}{4} & \frac{3}{4} \\
\end{bmatrix}
$
and 2 points are interpolated for each line.

Serving as the 2D representation to be located by $\mathcal{H}$ and $\mathcal{C}$, $\phi_g(\mathbf{x}_i)$ is defined to be the projected screen coordinates of $\tau(\mathbf{x}_i)$ given camera intrinsics $\text{K}_{3 \times 3}$ as
\begin{equation}
\phi_g(\mathbf{x}_i) = \text{K}_{3 \times 3} \tau(\mathbf{x}_i).
\end{equation}
$\phi_g(\mathbf{x}_i)$ implicitly encodes instance location on the image plane so that it is not an ill-posed problem to estimate egocentric pose from it directly. In summary, these IGRs can be computed with zero extra manual annotation, are easy to understand and contain rich information for estimating the instance orientation.

\section{Learning IGRs With Geometrical Prior}
\label{PI}
Sec.~\ref{MA} defines \emph{what} IGRs to learn, here we discuss \emph{how} to learn such representations more effectively. We propose to use geometrical prior knowledge for self-supervision and propose a mixed training strategy to take advantage of cheap unlabeled data for training. 
\begin{figure}[h]
	\begin{center}
		\includegraphics[width=0.65\linewidth, trim=5cm 5cm 5cm 5cm]{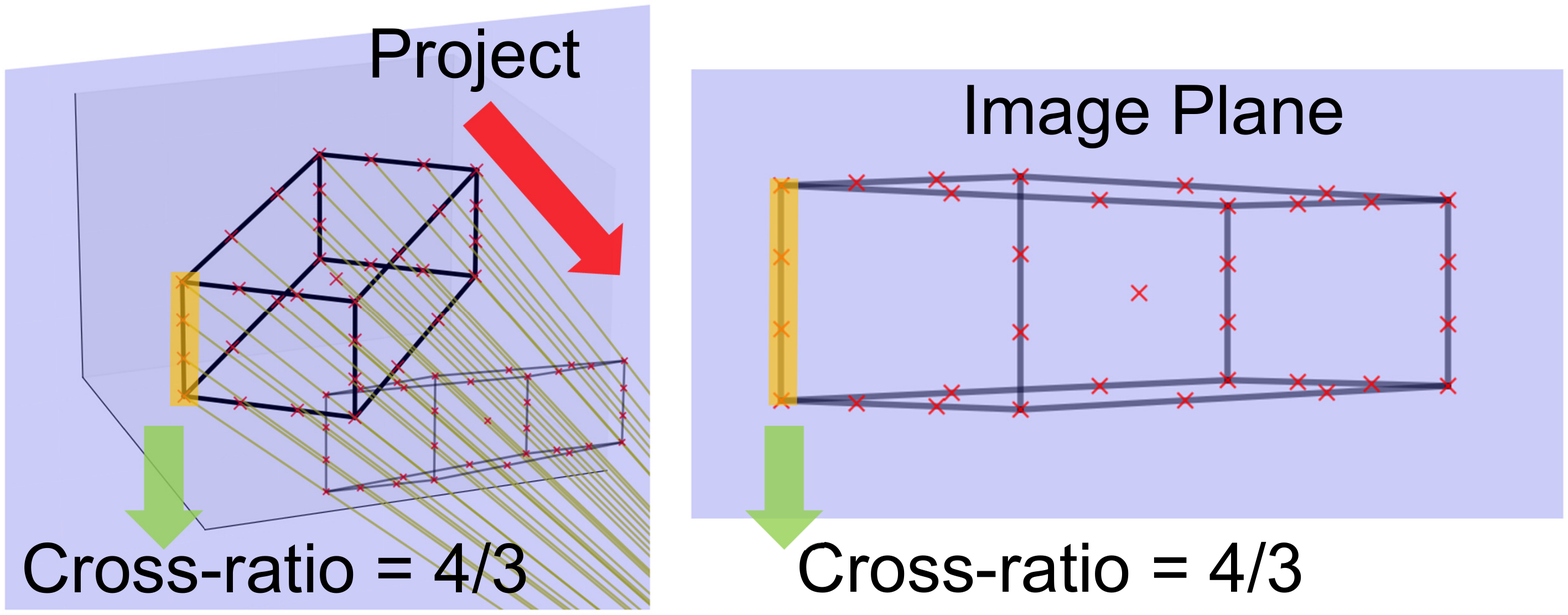}
	\end{center}
	\caption{Perspective projection of $\tau(\mathbf{x}_i)$ (left) and the zoomed image plane (right). Since the cross-ratio of every 4 collinear points is invariant after camera projection, the same value should be expected for a line on the screen as that of the corresponding line on the 3D bounding box.}
	\label{fig:cr}
\end{figure}

\subsection{Self-supervising by Preserving Cross-ratio}
The cross-ratio of 4 collinear points are invariant during perspective projection~\cite{hartley2003multiple}. Once $\psi(\mathbf{x}_i)$ is defined with a fixed interpolation matrix $B_{q \times 2}$, this ratio is fixed and invariant after perspective projection as shown in Fig.~\ref{fig:cr}. If the network $\mathcal{E}$ predicts a different value for such a ratio, one can penalize the network as supervisory signal. For every 4 points in $\phi_l(\mathbf{x}_i)$, we denote their ground truth 2D coordinates as $\phi_l(\mathbf{x}_i) = [\mathbf{v}_1, \mathbf{v}_2, \mathbf{v}_3, \mathbf{v}_4]$ and the ground truth cross-ratio as $cr = \frac{\|\mathbf{v}_3 - \mathbf{v}_1\|\|\mathbf{v}_4 - \mathbf{v}_2\|}{\|\mathbf{v}_3 - \mathbf{v}_2\|\|\mathbf{v}_4 - \mathbf{v}_1\|}$. Denote the predicted coordinates for such 4 points as $[\hat{\mathbf{v}}_1, \hat{\mathbf{v}}_2, \hat{\mathbf{v}}_3, \hat{\mathbf{v}}_4]$, we propose the cross-ratio loss function as 
\begin{equation}
L_{cr} = SmoothL1(cr^2 - \frac{\|\hat{\mathbf{v}}_3 - \hat{\mathbf{v}}_1\|^2\|\hat{\mathbf{v}}_4 - \hat{\mathbf{v}}_2\|^2}
	{\|\hat{\mathbf{v}}_3 - \hat{\mathbf{v}}_2\|^2\|\hat{\mathbf{v}}_4 - \hat{\mathbf{v}}_1\|^2}),
\end{equation}
where we choose $cr^2$ as the target because $\|\hat{\mathbf{v}}_p - \hat{\mathbf{v}}_q\|^2$ can be easily computed with vector inner product $<\hat{\mathbf{v}}_p - \hat{\mathbf{v}}_q, \hat{\mathbf{v}}_p - \hat{\mathbf{v}}_q>$ to avoid computing the square-root for $\|\hat{\mathbf{v}}_p - \hat{\mathbf{v}}_q\|$ .
\subsection{Utilizing Unlabeled Imagery}
\begin{figure}[h]
	\begin{center}
		\includegraphics[width=0.98\linewidth]{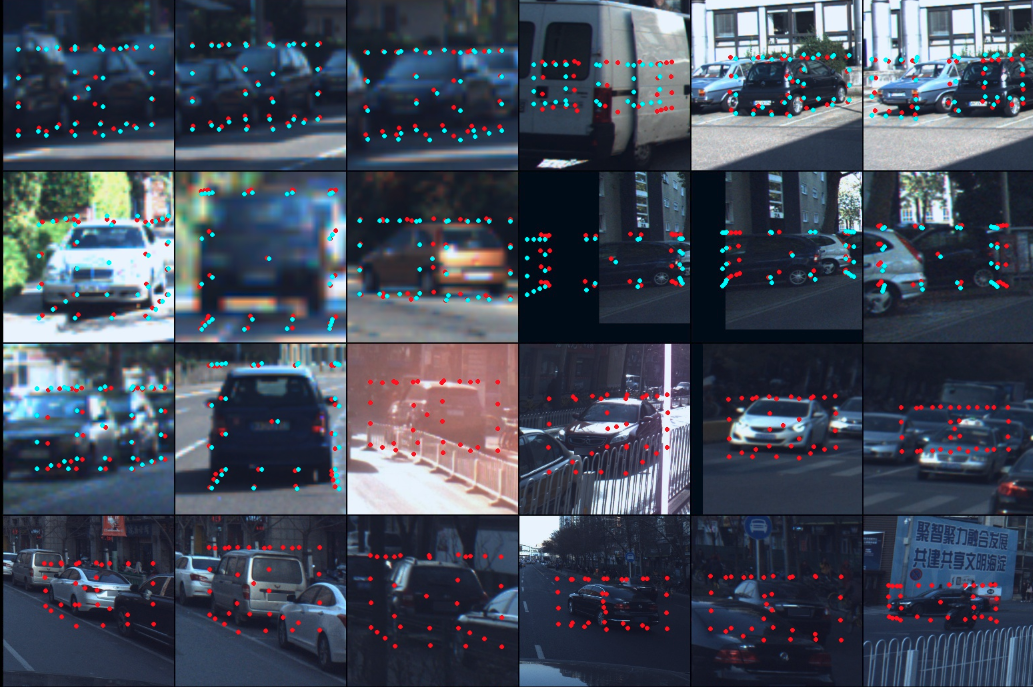}
	\end{center}
	\caption{A random training batch with ground truth $\phi_l(\mathbf{x})$ (blue) and prediction $\hat{\phi}_l(\mathbf{x})$ (red). By introducing the cross-ratio loss function, images with no labels can be employed into the training stage to regularize the learning of $\phi_l(\mathbf{x})$.}
\label{fig:mixture}
\end{figure}
Evaluating $L_{cr}$ does not require any ground truth information, which means an unlabeled vehicle instance can be used to supervise the learning of $\phi_l(\mathbf{x}_i)$. In addition, the preservation of cross-ratio holds irrespective of the camera intrinsics as long as a center projection model is used. This implies that imagery captured with different cameras at different locations can be used collectively and jointly. In our experiments, we crop unlabeled cars from~\cite{song2019apollocar3d} which were captured in a country different from those of KITTI~\cite{geiger2012we} and used cameras with different intrinsics. 

Ego-Net is supervised by hybrid types of loss functions and Algorithm~\ref{alg:loss} shows the forward-pass given a data batch consisting of both labeled and unlabeled instances. For the labeled ones, ground truth IGRs are available to evaluate heatmap loss $L_{hm}$ and 2D/3D representation losses $L_{2d}$ and $L_{3d}$. For the unlabeled images, only the proposed cross-ratio loss $L_{cr}$ is evaluated as an extra supervisory signal. Fig.~\ref{fig:mixture} shows a training batch containing both labeled and unlabeled cars. 

\begin{algorithm}
	\footnotesize
	\SetAlgoLined
	\KwIn{Labeled data with ground truth $D_{1} = \{\mathbf{x}_i, h(\mathbf{x}_i), \phi_l(\mathbf{x}_i), \psi(\mathbf{x}_i)\}_{i=1}^{N_1}$, unlabeled image inputs $D_{2} = \{\mathbf{x}_j\}_{i=1}^{N_2}$, model $\mathcal{E} = (\mathcal{H}, \mathcal{C}, \mathcal{L})$, loss functions for heatmaps, preserving cross-ratio, 2D and 3D representations as $L_{hm}, L_{cr}, L_{2d}$  and $L_{3d}$}
	\KwOut{Total loss $L_{total}$}
	\For{i=1:$N_1$}
	{$\hat{h}(\mathbf{x}_i) = \mathcal{H}(\mathbf{x}_i)$\; $\hat{\phi}_l(\mathbf{x}_i) = \mathcal{C}(\hat{h}(\mathbf{x}_i))$\; $\hat{\psi}(\mathbf{x}_i) = \mathcal{L}(\mathcal{A}_i(\hat{\phi}_l(\mathbf{x}_i)))$\;}
	\For{j=1:$N_2$}
	{$\hat{\phi}_l(\mathbf{x}_j) = \mathcal{C}(\mathcal{H}(\mathbf{x}_j))$\;}
	$L_{1} = \frac{1}{N_1}\sum_{i=1}^{N_1}L_{hm}(\hat{h}(\mathbf{x}_i), h(\mathbf{x}_i))$\;  
	$L_2 = \frac{1}{N_1}\sum_{i=1}^{N_1}L_{2d}(\hat{\phi}_l(\mathbf{x}_i), \phi_l(\mathbf{x}_i))$\; 
	$L_3 = \frac{1}{N_1}\sum_{i=1}^{N_1}L_{3d}(\hat{\psi}(\mathbf{x}_i), \psi(\mathbf{x}_i))$\;	 
	$L_4 = \frac{1}{N_1 + N_2}[\sum_{i=1}^{N_1}L_{cr}(\hat{\phi}_l(\mathbf{x}_i))
		+ \sum_{j=1}^{N_2}L_{cr}(\hat{\phi}_l(\mathbf{x}_j))]$\; 			
	$L_{total} = \sum_{i=1}^{4}L_i$	
\caption{Mixed training with unlabeled data}
\label{alg:loss}
\end{algorithm}

\begin{table*}[ht]
	\footnotesize
	\begin{center}
		\begin{tabular}{|c|c|c|c|c|c|c|c|}
			\hline
			Method &  Reference& Modality &\# of Viewpoints & Easy & Moderate & Hard & Average\\
			\cline{2-8}
			\hline
			\rowcolor{grayLight}
			Mono3D~\cite{chen2016monocular} & CVPR' 16   &RGB & Monocular &91.01&86.62 &76.84 &84.82\\
			\rowcolor{grayDark}
			Deep3DBox~\cite{mousavian20173d} & CVPR' 17  &RGB & Monocular&\nd{92.90}&\nd{88.75}&76.76 &\nd{86.14}\\
			\rowcolor{grayLight}
			ML-Fusion~\cite{xu2018multi} &CVPR' 18  &RGB &Monocular &90.35&87.03 & 76.37 &84.58\\			
			\rowcolor{grayDark}
			FQNet~\cite{liu2019deep} &CVPR' 19  &RGB &Monocular&92.58 &88.72 &76.85 &86.05\\
			\rowcolor{grayLight}
			GS3D~\cite{li2019gs3d} &CVPR' 19  &RGB &Monocular&85.79  &75.63  & 61.85 &74.42\\
			\rowcolor{grayDark}
			MonoPSR~\cite{ku2019monocular} &CVPR' 19  &RGB + LiDAR &Monocular&93.29 &	87.45 &	72.26 &84.33\\
			\rowcolor{grayLight}
			M3D-RPN~\cite{brazil2019m3d} &ICCV' 19  &RGB &Monocular&88.38 &82.81& 67.08 &79.42\\
			\rowcolor{grayDark}
			MonoPair~\cite{chen2020monopair} &CVPR' 20  &RGB &Monocular&91.65 &86.11 &76.45 &84.74\\	
			\rowcolor{grayLight}
			Disp R-CNN~\cite{sun2020disp} &CVPR' 20  &RGB + LiDAR &Stereo &93.02 &	81.70 &	67.16 &80.63\\				
			\rowcolor{grayDark}			
			DSGN~\cite{chen2020dsgn} &CVPR' 20  &RGB &Stereo&95.42&86.03& 78.27 &86.57\\	
			\rowcolor{grayLight}
			D4LCN~\cite{Ding_2020_CVPR} &CVPR' 20  &RGB &Monocular&90.01&82.08& 63.98 &78.69\\		
			\rowcolor{grayDark}
			RAR-Net~\cite{liu2020reinforced} &ECCV' 20  &RGB &Monocular&88.40&82.63 &66.90 &79.31\\	
			\rowcolor{grayLight}
			RTM3D~\cite{RTM3D} &ECCV' 20  &RGB &Monocular&91.75&86.73& \nd{77.18} &85.22\\								
			\rowcolor{grayDark}		
			Kinematic3D~\cite{kinematic-3d} & ECCV' 20  &RGB &Monocular&58.33 &	45.50 &	34.81 &46.21\\
			\hline
			\rowcolor{grayLight}
			Ego-Net (Ours)&CVPR' 21 &RGB &Monocular&\textbf{\te{\st{96.11}}{3.4}}&\textbf{\te{\st{91.23}}{2.8}}& \textbf{\te{\st{80.96}}{4.9}} &\textbf{\te{\st{89.43}}{3.8}} \\
			\hline
		\end{tabular}
	\end{center}
	\caption{System-level evaluation by comparing Average Orientation Similarity (AOS) with the SOTA learning-based methods for the KITTI test set in the car category. The second-highest performance among the single-view RGB-based approaches is shown in \nd{blue} and our improvement over it follows the $\uparrow$ signs. Without using LiDAR data~\cite{ku2019monocular} or temporal information~\cite{kinematic-3d}, our system out-performs previous monocular RGB-based methods and even stereo methods~\cite{chen2020dsgn, sun2020disp}.}	
	\label{tab:orientation}
\end{table*}

\section{Experiments}
\label{EX}
We first introduce the evaluation benchmark, followed by an ablation study to demonstrate the effectiveness of IGRs and geometrical self-supervision. Finally we compare the overall system performance with recent SOTA works and demonstrate how Ego-Net can improve other 3D object detection systems.
\subsection{Datasets and Evaluation Metrics}
\noindent\textbf{Datasets.} We employ the KITTI object detection benchmark~\cite{geiger2012we} for evaluation, which contains RGB images captured in outdoor scenes. The dataset is split into 7,481 training images and 7,518 testing images. The ground truth of the testing images is not released and thus all predictions were sent to the official server for evaluation. The training images are further split into a train set and a validation set containing 3,682 and 3,799 images respectively. Our Ego-Net is trained using only the train set and the validation set is used for hyper-parameter tuning. To obtain unlabeled car instances for training, we randomly crop images from ApolloCar3D~\cite{song2019apollocar3d} dataset.

\noindent\textbf{Evaluation Metrics.} We use the standard metric, \emph{Average Orientation Similarity} (AOS), to assess the overall system performance for joint vehicle detection and pose estimation. AOS is defined as $AOS = \frac{1}{41}\sum_{r\in\{0, 0.025, \dots, 1\}}\text{max}_{\tilde{r}:\tilde{r}\geq r}s(\tilde{r})$ where $r$ is the detection recall and $s(r) \in [0, 1]$ is the orientation similarity (OS) at recall level $r$. OS is defined as $s(r) = \frac{1}{|\mathcal{D}(r)|}\sum_{i \in \mathcal{D}(r)}\frac{1 + \text{cos}\Delta^{(i)}_{\theta}}{2}\delta_{i}$, where $\mathcal{D}(r)$ denotes the set of all object detections at recall rate $r$ and $\Delta^{(i)}_{\theta}$ is the difference in angle between estimated and ground truth vehicle pose of detection $i$. For the AOS evaluation, vehicle instances are grouped based on three difficulty levels: easy, moderate and hard. A smaller instance with more severe occlusion and truncation is considered to be more difficult.
\begin{figure*}[t]
	\begin{center}
		\includegraphics[width=1\linewidth, trim=0cm 1cm 0cm 1cm]{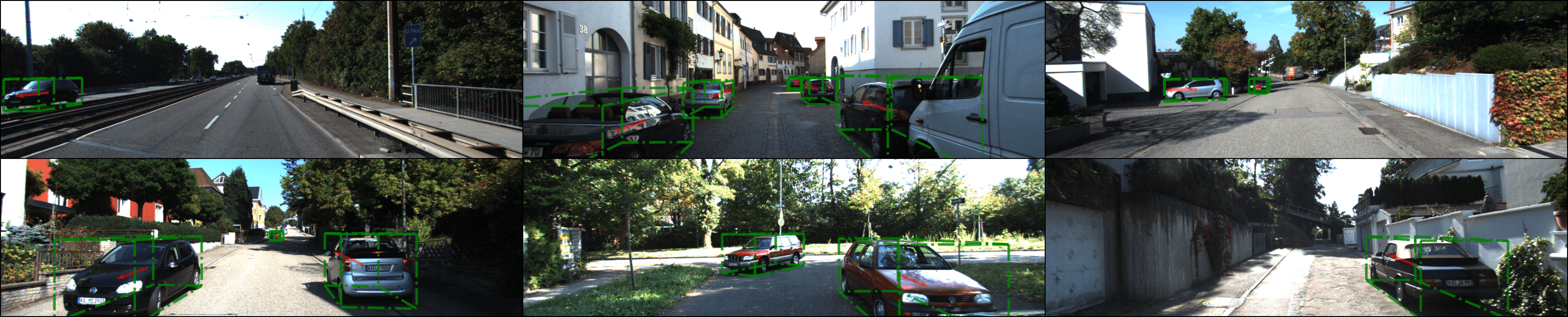}
	\end{center}
	\caption{Qualitative results on the KITTI validation set where the red arrows indicate the vehicle poses. More examples can be found in our supplementary material.}
	\label{fig:qualitative}
\end{figure*}
\subsection{Implementation Details} 
A detected instance is cropped and resized to a 256 by 256 patch as input to Ego-Net. HRNet-w48~\cite{sun2019deep, wang2020deep} is used as the fully-convolutional sub-model $\mathcal{H}$ to regress heatmaps of size 64 by 64. $L2$ loss is used for $L_{hm}$ and $L_{3d}$, while $L1$ loss is used for $L_{2d}$. More training details are included in our supplementary material.

\subsection{Ablation Study}
\noindent \textbf{Direct regression vs. learning IGRs} To evaluate the effectiveness of our proposed IGRs, we compare the pose estimation accuracy with learned IGRs to that of a baseline which directly regresses pose angles from the instance feature maps. AOS is influenced by the recall of car detection and is upper-bounded by 2D Average Precision ($AP_{2D}$) of the detector~\cite{geiger2012we}. To eliminate the impact of the used object detector, we compare AOS on all annotated vehicle instances in the validation set. This is equivalent to measuring AOS with $AP_{2D}=1$ so that the orientation estimation accuracy becomes the only focus. The comparison is summarized in Tab.~\ref{tab:ablation1}. Note learning IGRs outperforms the baseline by a significant margin. Deep3DBox~\cite{mousavian20173d} is another popular architecture\footnote{https://github.com/smallcorgi/3D-Deepbox} that performs direct angle regression, but utilizes a complex \emph{MultiBin} loss function for joint classification and regression. Without designing such loss functions, our approach outperforms it with the novel design of IGRs and does not need discrete pose classification. Qualitative visualization of pose predictions on the validation set is shown in Fig.~\ref{fig:qualitative}.

\newcommand{\TableEntry}[2]{$\text{#1}_{\downarrow \text{#2}\%}$}
\begin{table}[h]
	\footnotesize
	\centering
	\begin{tabular}{|l|c|c|c|}
		\hline
		\multirow{2}{*}{Method} & \multicolumn{3}{c|}{$AOS$ when $AP_{2D}=100.00$}\\ \cline{2-4}
		&  Easy & Moderate & Hard \\ 
		\hline
		\rowcolor{grayDark}
		B  &  95.22 &92.14  &88.37 \\
		\rowcolor{grayLight}
	    Deep3DBox~\cite{mousavian20173d}  &98.48 &95.81  &91.98 \\
		\rowcolor{grayDark}
		B+IGRs (Ego-Net)  &  \te{99.58}{1.1} &\te{99.06}{3.4}  &\te{96.55}{5.0}\\	
		\hline
	\end{tabular}
	\caption{Module-level evaluation assuming perfect object detection ($AP_{2D}=100.00$) in the KITTI validation set. B:~baseline using direct regression. +IGR:~adding intermediate geometric representations. Improvements compared with \cite{mousavian20173d} follow the $\uparrow$ signs.}
	\label{tab:ablation1}
\end{table} 

\noindent \textbf{Effect of the mixed training strategy}. To study whether the introduction of $L_{cr}$ and unlabeled data can improve the learning of $\phi_g(\mathbf{x}_i)$ and result in improved accuracy of locating $\phi_g(\mathbf{x}_i)$, we re-train Ego-Net without using the cross-ratio component and unlabeled data. The quantitative comparison is shown in Tab.~\ref{tab:cru}. \emph{Percentage of Correct Keypoints} (PCK@$X$) gives the ratio of correctly located joints that have pixel error smaller than $X$ times a pre-defined threshold. This threshold is different for each instance and depends on the instance size so that PCK is not influenced by car size variation. We define the threshold as one third of the bounding box height for each instance. Tab.~\ref{tab:cru} shows that the accuracy of locating screen coordinates improves after adding unlabeled data for training. Despite the unlabeled images do not have any 3D annotation and were taken in a different country with different camera intrinsics, they benefit model generalization. We believe the reason is that the introduction of these unlabeled images mitigates the appearance bias towards the training data. 

\begin{table}[h]
	\footnotesize
	\centering
	\begin{tabular}{|l|c|c|c|}
		\hline
		Method &  PCK@0.1 & PCK@0.2 & PCK@0.3 \\
		\hline
		\rowcolor{grayDark}
		w/o mixed training  & 31.3  &64.1 &79.3 \\
		\rowcolor{grayLight}
		w/ mixed training& 31.9  &64.8 & 79.6\\	
		\hline	
	\end{tabular}
	\caption{Comparison of coordinate localization performance on KITTI validation set with and without using unlabeled data for mixed training. Higher PCK is better.}
	\label{tab:cru}
\end{table} 
	
\subsection{Comparison with SOTA works} 
\noindent \textbf{Joint vehicle detection and pose estimation} performance is measured by AOS. Tab.~\ref{tab:orientation} compares the AOS of our system using Ego-Net with other SOTA approaches on KITTI test set. Among the single-view RGB-based approaches, Ego-Net outperforms all other deep models by a clear margin. Ego-Net even outperforms DSGN~\cite{chen2020dsgn} and Disp R-CNN~\cite{sun2020disp} that use stereo cameras. In addition, our approach using a single image outperforms Kinematic3D~\cite{kinematic-3d} which exploits temporal information using RGB video. This result indicates that vehicle pose can be reliably estimated with a single frame. In fact, our approach can also serve for per-frame initialization of temporal models.

\begin{table}[h]
	\footnotesize
	\centering
	\begin{tabular}{|l|c|c|c|}
		\hline
		\multirow{3}{*}{Method} & \multicolumn{3}{c|}{$AOS\&AP_{2D}$,~~$AOS \le AP_{2D}$}\\ \cline{2-4}
		 &  Easy & Moderate & Hard \\ \cline{2-4}
			& \multicolumn{3}{c|}{$AP_{2D}$}\\ 
		\hline
		\rowcolor{grayDark}
		M3D-RPN~\cite{brazil2019m3d}  &  90.28 & 83.75  & 67.72 \\
		\rowcolor{grayLight}
		D4LCN~\cite{Ding_2020_CVPR}  & 92.80 & 84.43  & 67.89 \\
		\hline
		Method	& \multicolumn{3}{c|}{$AOS$}\\ 
		\hline
		\rowcolor{grayDark}
		M3D-RPN~\cite{brazil2019m3d}  &  88.79 & 81.25  & 65.37 \\
		\rowcolor{grayLight}
		M3D-RPN + Ego-Net  &  \te{90.20}{1.6} &\te{83.60}{2.9}  & \te{67.53}{3.3} \\
		\rowcolor{grayDark}
		D4LCN~\cite{Ding_2020_CVPR}  & 91.74  &82.96  &66.45\\
		\rowcolor{grayLight}
		D4LCN + OCM3D~\cite{peng2021ocm3d} &  92.12 &83.27 &66.81 \\	
		\rowcolor{grayDark}		
		D4LCN + Ego-Net  &  \te{92.62}{1.0} &\te{84.25}{1.6}  &\te{67.60}{1.7} \\	
		\hline	
	\end{tabular}
	\caption{AOS evaluation on KITTI validation set. After employing Ego-Net, the vehicle pose estimation accuracy of SOTA 3D object detection systems can be improved. The space for AOS improvement is upper-bounded by $AP_{2D}$.}
	\label{tab:module}
\end{table} 
\begin{table}[h]
	\footnotesize
	\centering
	\begin{tabular}{|l|c|c|c|}
		\hline
		\multirow{2}{*}{Method} & \multicolumn{3}{c|}{$AP_{BEV}$}\\ \cline{2-4}
		&  Easy & Moderate & Hard \\ 
		\hline
		\rowcolor{grayDark}
		ROI-10D~\cite{manhardt2019roi}  &  14.04 &3.69  &3.56 \\
		\rowcolor{grayLight}
		Mono3D++~\cite{he2019mono3d++}  &  16.70 &11.50  &10.10 \\
		\rowcolor{grayDark}
		MonoDIS~\cite{simonelli2019disentangling}  &  24.26 &18.43  &16.95 \\
		\rowcolor{grayLight}
		D4LCN~\cite{Ding_2020_CVPR}   &31.53 &22.58  &17.87 \\
		\rowcolor{grayDark}		
		D4LCN + Ego-Net (Ours)  &  \textbf{33.60} &\textbf{25.38} &\textbf{22.80} \\	
		\hline
	\end{tabular}
	\caption{$AP_{BEV}$ evaluated on KITTI validation set. Ego-Net can correct the erroneous pose predictions from~\cite{Ding_2020_CVPR} as shown in Fig.~\ref{fig:quali_comp}.}
	\label{tab:bev}
\end{table} 

\begin{figure}
	\begin{center}
		\includegraphics[width=1\linewidth, trim=1cm 3cm 1cm 1cm]{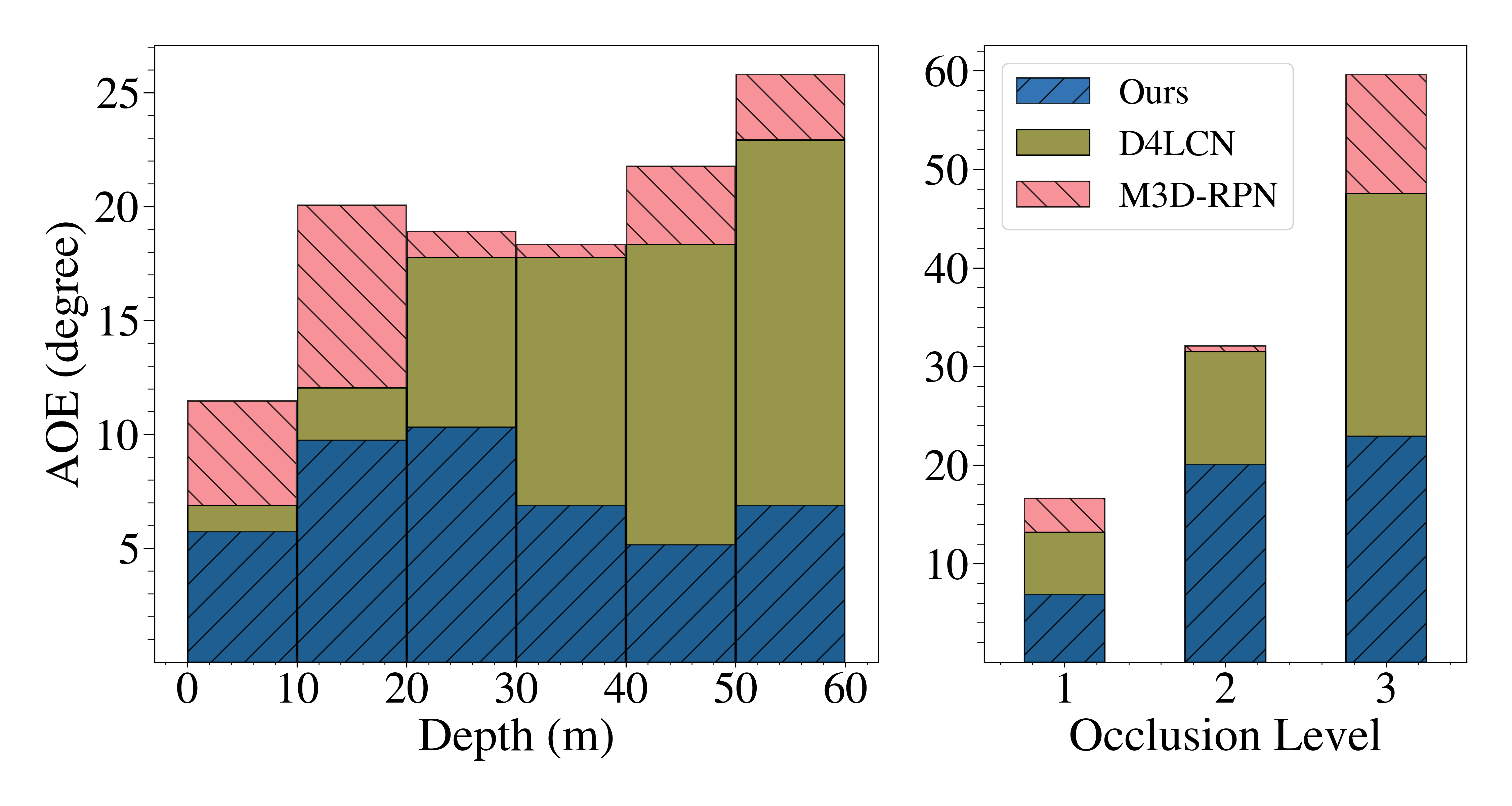}
	\end{center}
	\caption{Average orientation error (AOE) on KITTI validation set in different depth ranges and occlusion levels. Our approach is robust to distant and partially occluded instances.}
	\label{fig:aoe}
\end{figure}

\noindent \textbf{Ego-Net as a module} can be used as the pose estimation branch of other 3D object detection systems. To assess if Ego-Net can help improve the pose estimation accuracy of other 3D object detection systems, compared to their original pose estimation branch, we fuse Ego-Net with some recent works with open-source implementations. We took the detected instances produced by other approaches and compared the estimated poses by Ego-Net with those produced by their original pose estimation modules. The result is summarized in Tab.~\ref{tab:module}. While AOS depends on the detection performance of these methods, fusing Ego-Net consistently improves the pose estimation accuracy of these approaches. This indicates that Ego-Net is robust despite the performance of a vehicle detector varies with different recall levels. For these detected instances (true positives) we plot the distribution of orientation estimation error versus different depth ranges and occlusion levels\footnote{The occlusion levels are annotated for each instance in KITTI.} in Fig.~\ref{fig:aoe}. The error in each bin is the averaged egocentric pose error for those instances that fall into it. While the performance of M3D-RPN~\cite{brazil2019m3d} and D4LCN~\cite{Ding_2020_CVPR} degrades significantly for distant and occluded cars, the errors of our approach increase gracefully. We believe that explicitly learning the vehicle parts makes our model more robust to occlusion as the visible parts can provide rich information for pose estimation.

\begin{figure}
	\begin{center}
		\includegraphics[width=1\linewidth]{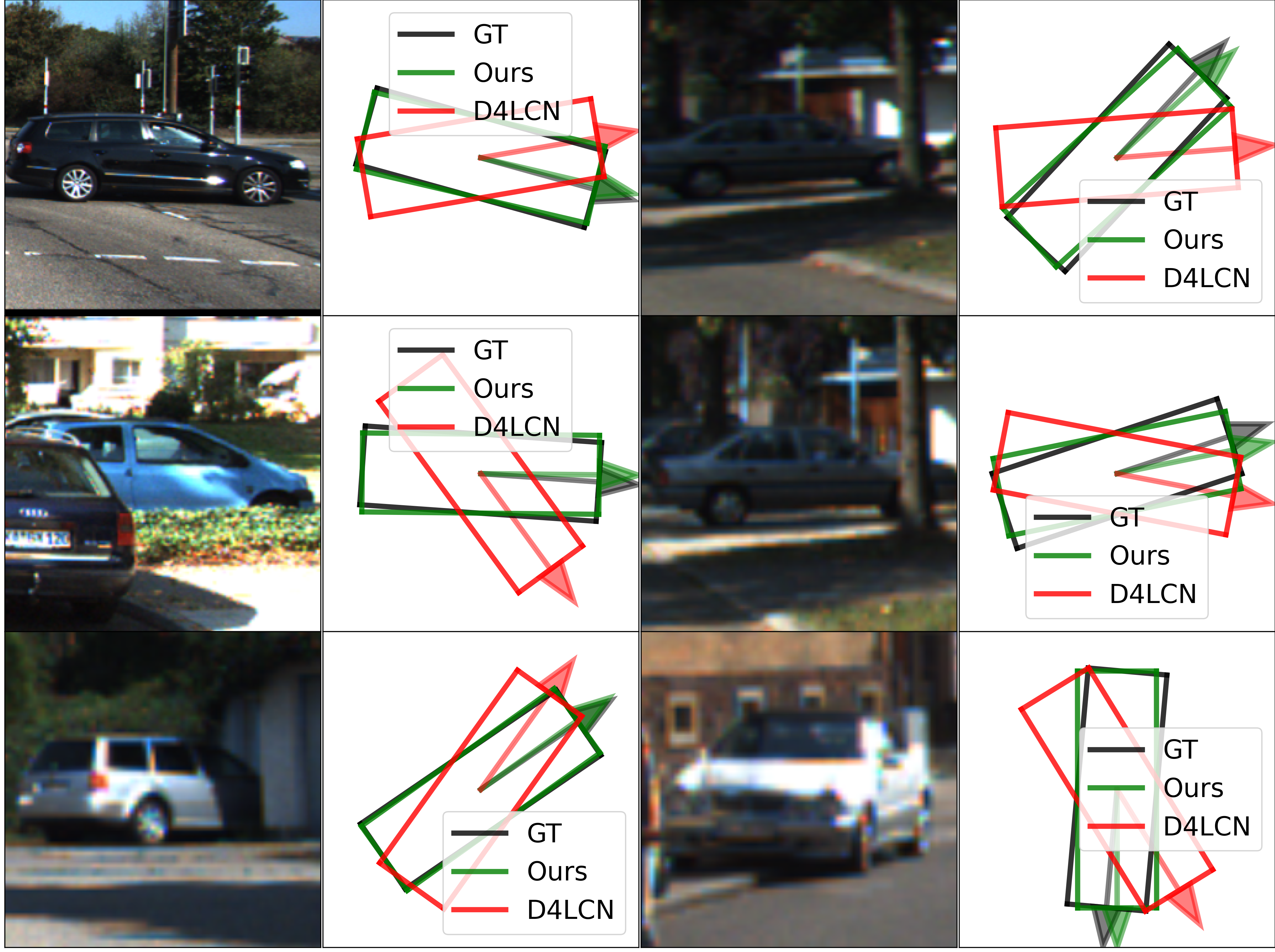}
	\end{center}
	\caption{Detected instances in KITTI validation set along with the comparison of the predicted egocentric poses in bird's eye view.}
	\label{fig:quali_comp}
\end{figure}

\noindent \textbf{Bird's eye view (BEV) evaluation} compared with D4LCN~\cite{Ding_2020_CVPR} is visualized in Fig.~\ref{fig:quali_comp}. The egocentric pose predictions are shown under bird's eye view and the arrows point to the heading direction of those vehicles. From the local appearance it can be hard to tell whether certain cars head towards or away from the camera, as validated by the erroneous predictions of D4LCN~\cite{Ding_2020_CVPR} since it regresses pose from local features. In comparison, our approach gives accurate egocentric pose predictions for these instances and others that are partially occluded. Quantitative comparison with several 3D object detection systems~on the KITTI validation set is shown in Tab.~\ref{tab:bev}. Note that utilizing Ego-Net can correct wrong pose predictions especially for difficult instances, which leads to improved 3D IoU and results in significantly improved $AP_{BEV}$.
	
\section{Conclusion} 
We propose a new deep model for estimating 3D vehicle pose in the camera coordinate system. The heatmaps of object parts are first extracted from local object appearance, which are mapped to the screen coordinates and then further lifted to a 3D object pose. A novel loss function that encourages the preservation of cross-ratio is proposed to improve the estimation accuracy of the screen coordinates. Based on the definition of intermediate representations and the cross-ratio loss, unlabeled instances can be used jointly with labeled objects to improve representation learning. Our system hits a record high for the vehicle detection and pose estimation task on KITTI among the monocular RGB-based deep models. 

Several future directions remain to be explored. For example, the definition of intermediate representations could consider instance point-clouds if LiDAR data is available. More geometrical prior knowledge apart from preserving cross-ratio can be incorporated. Apart from unlabeled instances, graphics-rendered cars can also be incorporated whose 3D ground truth are known.

\noindent \textbf{Acknowledgments} This work was partially supported by Hong Kong Research Grants Council (RGC) General Research Fund (GRF) 16203319. We acknowledge the support
of NVIDIA Corporation with the donation of one Titan Xp GPU used for this research.

{\small
	\bibliographystyle{ieee_fullname}
	\bibliography{egbib}
}

\setcounter{section}{0}
\clearpage
\begin{center}
	\large{\textbf{Supplementary Material}}
\end{center}
The supplementary materials are organized in the following sections:
\begin{itemize}
	\item \hyperref[sec1]{\textbf{Section 1}} details the \textbf{network architecture and hyper-parameters} used in the experiments.
	\item \hyperref[sec2]{\textbf{Section 2}} includes \textbf{training details} such as data pre-processing and relevant parameters.
	\item \hyperref[sec3]{\textbf{Section 3}} gives \textbf{extended results} on KITTI validation set.
\end{itemize}
\section{Architecture Hyper-parameters} \label{sec1}
\subsection{Ego-Net}
The detailed network hyper-parameters of Ego-Net is shown in Tab.~\ref{tab:arch}. We use HRNet-w48~\cite{sun2019deep} as the fully-convolutional model, which can be replaced by other CNNs as well. After regressing heatmaps of size 66 by 64 by 33. We add two channels by appending xy maps to help coordinate regression. ResBlock-conv is a residual block whose residual branch is composed of two consecutive 3 by 3 convolution layers. The input branch is a convolution layer with kernel size 1 and stride 2 to down-sample the input so that it can be added with the output of the residual branch. ResBlock-fc is a residual block whose residual branch is composed of two fully-connected layers. After each fully-connected or convolution layer, batch normalization and ReLU activation are applied. Dropout layers with dropout rate 0.5 are appended after the fully-connected layers.

Finally, to obtain egocentric pose as angle from the predicted 3D representation $\psi(\mathbf{x}_i)$, one can simply extract the directions from three orthogonal edges of the predicted interpolated cuboid. In implementation we estimate a rigid transformation from the predicted cuboid to a template cuboid at canonical pose using OpenCV.
 
\subsection{Baseline}
The architecture of the baseline model used in the ablation study is shown in Tab.~\ref{tab:baseline}. The feature maps are reduced in spatial size with several residual blocks and the regression target is the angle represented on the unit circle, i.e., a two-vector $[cos(\theta), sin(\theta)]$ following previous works~\cite{hara2017designing, mousavian20173d}.
\subsection{Using Ego-Net in two-stage architecture}
Combining $\mathcal{E}$ with another translation estimation module $\mathcal{T}$ is straightforward. The pseudo code in Alg.~\ref{alg:example} shows the example when using Ego-Net with M3D-RPN~\cite{brazil2019m3d}. For each 2D bounding box $b_i$ predicted by M3D-RPN, the local appearance is cropped and fed to Ego-Net to obtain its egocentric pose $\theta_i$, while the remaining predictions such as translation and dimension predicted by M3D-RPN can be unchanged. In fact, instance dimension is also encoded by the predictions of Ego-Net so that the user can use Ego-Net for dimension estimation as well.

\begin{algorithm}
	\footnotesize
	\SetAlgoLined
	\KwIn{M3D-RPN $\mathcal{T}$, Ego-Net $\mathcal{E}$, input image I}
	\KwOut{Detected instances with accurate egocentric pose $(b_i, w_i, h_i, l_i, x_i, y_i, z_i, \theta_i)_{i=1}^N$}
	$(b_i, w_i, h_i, l_i, x_i, y_i, z_i)_{i=1}^N$ = $\mathcal{T}(I)$\;
	\For{i=1:$N$}{$\theta_i= \mathcal{E}(Crop(I, b_i))$} 
	\caption{Combine Ego-Net with M3D-RPN to form $\mathcal{M}$ } 
	\label{alg:example}
\end{algorithm}

\newcommand{\blocka}[2]{\multirow{3}{*}{\(\left[\begin{array}{c}\text{3$\times$3, #1}\\[-.1em] \text{3$\times$3, #1} \end{array}\right]\)$\times$#2}
}
\newcommand{\blockb}[3]{\multirow{3}{*}{\(\left[\begin{array}{c}\text{1$\times$1, #2}\\[-.1em] \text{3$\times$3, #2}\\[-.1em] \text{1$\times$1, #1}\end{array}\right]\)$\times$#3}
}
\begin{table*}[h]
	\footnotesize
	\begin{center}
		\resizebox{0.7\linewidth}{!}{
			\begin{tabular}{c|c|c|c}
				\hline
				Layer/sub-model name & Type & Input size & Output size  \\
				\hline
				FCN $\mathcal{H}$ &  HRNet-w48~\cite{sun2019deep} & 256$\times$256$\times$3  & 64$\times$64$\times$33 \\
				\hline
				$\mathcal{C}$-1&  ResBlock-conv, stride 2& 64$\times$64$\times$35 & 32$\times$32$\times$66  \\
				$\mathcal{C}$-2&  ResBlock-conv, stride 2& 32$\times$32$\times$66 & 16$\times$16$\times$66  \\
				$\mathcal{C}$-3&  ResBlock-conv, stride 2& 16$\times$16$\times$66 & 8$\times$8$\times$66  \\				
				$\mathcal{C}$-4&  ResBlock-conv, stride 2& 8$\times$8$\times$66 & 4$\times$4$\times$66  \\
				$\mathcal{C}$-5&  Conv4$\times$4, stride 1& 4$\times$4$\times$66 & 1$\times$1$\times$66 \\				
				\hline
				$\mathcal{L}$-1&  FC& 1$\times$66 & 1$\times$1024  \\
				$\mathcal{L}$-2&  ResBlock-fc& 1$\times$1024  & 1$\times$1024  \\
				$\mathcal{L}$-3&  ResBlock-fc& 1$\times$1024  &1$\times$1024  \\				
				$\mathcal{L}$-4&  FC& 1$\times$1024  &1$\times$96  \\					
				\hline
			\end{tabular}
		}
	\end{center}
	\caption{Detailed architecture of Ego-Net. FC is a fully-connected layer. See the text for the explanation of ResBlock.}
	\label{tab:arch}
\end{table*}
\begin{table*}[h]
	\footnotesize
	\begin{center}
		\resizebox{0.7\linewidth}{!}{
			\begin{tabular}{c|c|c|c}
				\hline
				Layer/sub-model name & Type & Input size & Output size  \\
				\hline
				FCN $\mathcal{H}$ &  HRNet-w48~\cite{sun2019deep} & 256$\times$256$\times$3  & 64$\times$64$\times$48 \\
				\hline
				$\mathcal{B}$-1&  Conv1$\times$1, stride 1& 64$\times$64$\times$48 & 64$\times$64$\times$256  \\
				$\mathcal{B}$-2&  ResBlock-conv, stride 2& 64$\times$64$\times$256 & 32$\times$32$\times$256  \\
				$\mathcal{B}$-3&  ResBlock-conv, stride 2& 32$\times$32$\times$256 & 16$\times$16$\times$256  \\				
				$\mathcal{B}$-4&  ResBlock-conv, stride 2& 16$\times$16$\times$256 & 8$\times$8$\times$256  \\
				$\mathcal{B}$-5&  ResBlock-conv, stride 2& 8$\times$8$\times$256 & 4$\times$4$\times$256 \\				
				$\mathcal{B}$-6&  Average Pooling 2D, kernel size 4& 4$\times$4$\times$256 & 1$\times$1$\times$256  \\
				$\mathcal{B}$-7&  FC& 1$\times$256  & 1$\times$256  \\
				$\mathcal{B}$-8&  FC& 1$\times$256 &1$\times$2  \\								
				\hline
			\end{tabular}
		}
	\end{center}
	\caption{Detailed architecture of the baseline model used in the ablation study.}
	\label{tab:baseline}
\end{table*}
\section{Training Details} \label{sec2}
\subsection{Hyper-parameters for training}
$\mathcal{H}$ and $\mathcal{C}$ are trained together to perform the localization of screen coordinates of the projected interpolated cuboid. During training, each instance is cropped from the image given a 2D bounding box. This bounding box is determined by first computing the tight bounding box enclosing $\phi_g(\mathbf{x}_i)$ and then extend it to get an aspect ratio of 1. To enhance the robustness of the model against object detector, we randomly translate and scale the bounding box during training. The cropped patch has spatial size 256 by 256 after an affine warping $\mathcal{A}_i$ (actually a similarity transformation with only scaling and translation). The ground truth local coordinates $\phi_l(\mathbf{x}_i)$ are obtained by applying $\mathcal{A}_i$ to the global coordinates. There are some annotated instances in KITTI~\cite{geiger2012we} train set that lie out of the image plane. We remove such outliers by inspecting the global coordinates $\phi_g(\mathbf{x}_i)$ of the training instances. Those instances that have more than 30 percent key-points outside of the image plane are discarded. Ground truth heatmaps and coordinates are prepared based on the ground truth local coordinates to evaluate $L_{hm}$ and $L_{2d}$. The ground truth heatmaps are of size 64 by 64 and obtained by drawing Gaussian dots at the ground truth locations where the sigma is 1 pixel. The ground truth local coordinates are normalized to the input size to be in the range of 0 to 1. We use Adam optimizer to train the network for 50 epochs with a batch size of 24 images. The start learning rate is 0.001, which is multiplied by 0.5 after every 10 epochs. For each image, if the total instance count is smaller than 12, we add unlabeled car instances that are cropped from Apollo3DCar~\cite{song2019apollocar3d} dataset until 12 instances are present. 

$\mathcal{L}$ is trained with ground truth 2D-3D pairs $(\phi_g(\mathbf{x}_i), \psi(\mathbf{x}_i))$ to lift the located coordinates into a 3D cuboid. The training data is prepared by projecting the ground truth 3D cuboid into 2D image plane with the provided camera intrinsics in KITTI. From the 3,682 training images in the KITTI, there are 13.6k 2D-3D pairs. We augment the data by randomly rotate the ground truth 3D cuboid to generate 100 times more training pairs so that the final training set size is 1.37M. We use Adam optimizer to train the network with $L_{3d}$ for 300 epochs. The start learning rate is 0.001, which is multiplied by 0.5 after every 50 epochs.

\subsection{Cross-ratio loss}
\begin{figure}[h]
	\begin{center}
		\includegraphics[width=0.95\linewidth, trim=0cm 1.5cm 0cm 1cm]{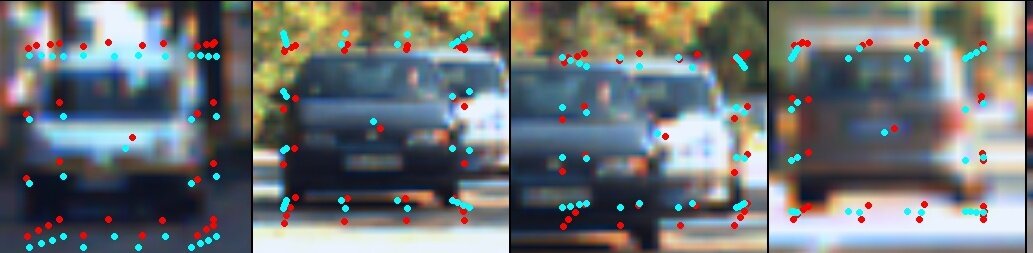}
	\end{center}
	\caption{Fore-shortened edges are not used during computing the cross-ratio loss function.}
	\label{fig:cr2}
\end{figure}

Depending on the vehicle pose, certain edges of a projected 3D bounding box may appear very short due to the fore-shortening effect of perspective projection. See Fig.~\ref{fig:cr2} for an example, the second vehicle faces towards the camera and the four parallel edges (along the heading direction) of its 3D bounding box appear to collapse into four points. When computing the cross-ratio of collinear points that are very close to each other, the denominator can be small and produce unstable gradient in our experiment. Thus we decide to discount the fore-shortened edges during computing $L_{cr}$. In implementation, we inspect every group of four collinear points and give up computing $L_{cr}$ if the minimal pair-wise distance of the four points ($d_m$) is smaller than a threshold $D$. This step can be illustrated as
\begin{align*}
L_{cr} & = \begin{cases}
SL1(cr^2 - \frac{\|\hat{\mathbf{v}}_3 - \hat{\mathbf{v}}_1\|^2\|\hat{\mathbf{v}}_4 - \hat{\mathbf{v}}_2\|^2}
{\|\hat{\mathbf{v}}_3 - \hat{\mathbf{v}}_2\|^2\|\hat{\mathbf{v}}_4 - \hat{\mathbf{v}}_1\|^2}), &\mbox{if } d_m > D,\\
0, & \mbox{else}.
\end{cases}
\end{align*}
We use $D$=0.15 in the experiments. In addition, the definition of the cross-ratio loss is based on four collinear points and is not applicable if the four points are not collinear. We disable $L_{cr}$ when the training starts and switch $L_{cr}$ on after the model runs one epoch on KITTI so that the predicted points will be almost collinear.

\begin{figure*}[t]
	\begin{center}
		\includegraphics[width=0.95\linewidth, trim=0cm 1cm 0cm 1cm]{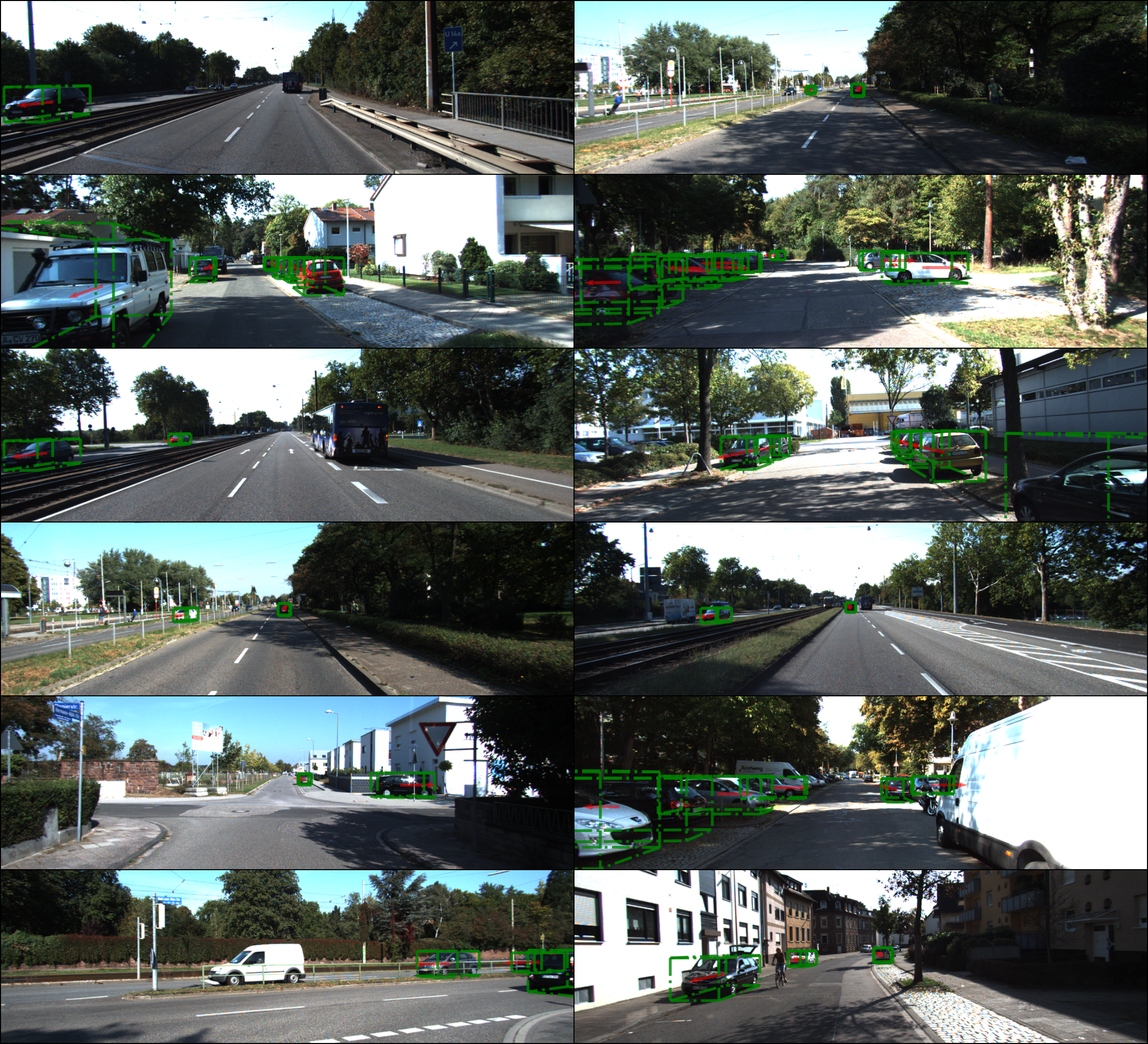}
	\end{center}
	\caption{Qualitative results on the KITTI validation set where the red arrows indicate the vehicle poses.}
	\label{fig:qualitative_supp}
\end{figure*}

\subsection{Loss balancing}
During training the implemented loss functions $L_{hm}$, $L_{2d}$ and $L_{cr}$ return value of different scales. The ground truth heatmaps are sparse since the Gaussian dots only occupy a small region, resulting in small $L_{hm}$ in mean square loss computation. We adjust the scale of these loss functions empirically with a weight of 1, 0.1 and 0.005 respectively. 

\section{Extended Validation Results} \label{sec3}
\subsection{Quantitative results}
Tab.~\ref{tab:cru} in the main text compared the accuracy of locating screen coordinates. Here such results are extended to show the influence on the 3D pose estimation accuracy. As summarized in Tab.~\ref{tab:cru2}, adding cross-ratio loss and unlabeled data also help improve the 3D pose estimation accuracy for more difficult cases.

\begin{table}[h]
	\centering
	\begin{tabular}{|l|c|c|c|}
		\hline
		Method &  Easy & Medium & Hard \\
		\hline
		\rowcolor{grayDark}
		w/o mixed training  &  99.75 &98.95  &96.48 \\
		\rowcolor{grayLight}
		w/ mixed training &99.58  &99.06  &96.55\\	
		\hline
	\end{tabular}
	\caption{AOS evaluated for all vehicle instances on KITTI validation set ($AOS$ when $AP_{2D}=100.00$) with and without using unlabeled data for mixed training.}
	\label{tab:cru2}
\end{table} 

\subsection{Qualitative results}

More qualitative results on KITTI validation set such as Fig.~\ref{fig:qualitative_supp} can be downloaded at this https \href{https://drive.google.com/drive/folders/1pHDkVcAxeK2QGZuSkz9qIGcEMgsty-oU?usp=sharing}{URL}.
\end{document}